# An Anthological Review of Research Utilizing MontyLingua, a Python-Based End-to-End Text Processor

Maurice HT Ling
Department of Zoology, The University of Melbourne, Australia

Correspondence: mauriceling@acm.org

## Abstract
MontyLingua, an integral part of ConceptNet which is currently the largest commonsense knowledge base, is an English text processor developed using Python programming language in MIT Media Lab. The main feature of MontyLingua is the coverage for all aspects of English text processing from raw input text to semantic meanings and summary generation, yet each component in MontyLingua is loosely-coupled to each other at the architectural and code level, which enabled individual components to be used independently or substituted. However, there has been no review exploring the role of MontyLingua in recent research work utilizing it. This paper aims to review the use of and roles played by MontyLingua and its components in research work published in 19 articles between October 2004 and August 2006. We had observed a diversified use of MontyLingua in many different areas, both generic and domain-specific. Although the use of text summarizing component had not been observe, we are optimistic that it will have a crucial role in managing the current trend of information overload in future research.

## Categories and Subject Descriptors
H.5.2 [**User Interfaces**]: Natural Language
I.2.7 [**Natural Language Processing**]: Language Parsing

## 1. Introduction

MontyLingua (web.media.mit.edu/~hugo/montylingua/) is a natural language processing engine primarily developed by Hugo Liu in MIT Media Labs using the Python programming language, which is entitled as "*an end-to-end natural language processor with common sense*" (Liu, 2004). It is an entire suite of individual tools catering to all aspects of English text processing, ranging from raw text to the extraction of semantic meanings and summary generation; thus, end-to-end. Commonsense is incorporated into MontyLingua's part-of-speech (POS) tagger, MontyTagger, as contextual rules.

MontyTagger was previously released by Hugo Liu as a standalone Brill-styled (Brill, 1995) POS tagger in 2002 but is now packaged with other components forming MontyLingua. A Java version of MontyLingua, built using Jython, had also been released. MontyLingua is also an integral part of ConceptNet (Liu and Singh, 2004), presently the largest commonsense knowledge base (Hsu and Chen, 2006), as a text processor and understander, as well as forming an application programming interface (API) to ConceptNet. At the same time, it had also been incorporated into Minorthird, a collection of Java classes for storing text, annotating text, and learning to extract entities and categorize text, written by William W. Cohen in Carnegie Mellon University (Cohen, 2004).

To date, there were only 2 modules specifically written to process English text using Python: MontyLingua and NLTK (Loper and Bird, 2002). NLTK (Natural Language Toolkit) was developed by Edward Loper (University of Pennsylvania) and Steven Bird (The University of Melbourne) with the main purpose of teaching computational linguistics to computer science students (Loper and Bird, 2002). Thus, NLTK is more of a text processing library from which text processing engines, such as MontyLingua, could be developed from, rather than a suite of usable tools. This implied that MontyLingua could be re-implemented using NLTK but had not been done. Another popular text processor is GATE (Cunningham, 2000), which was developed in Java. The main difference between GATE and MontyLingua is that GATE is a template processing engine rather than natural language processing.

ConceptNet and MontyLingua, as well as 15 applications of ConceptNet, had been previously been described (Liu and Singh, 2004). However, there has not been any review since October 2004 updating the state-of-the-art use of either ConceptNet or MontyLingua. At the same time, there has not been any review examining the roles played by MontyLingua and its components in recent research work, especially post-October 2004. This paper aims to review the use of and roles played by MontyLingua and its components in research work published between October 2004 and August 2006.

The rest of this paper is organized as follows: Section 2 describes the distinctive feature and main components of MontyLingua. In Section 3, we review 23 research publications, that were published between October 2004 and August 2006, for the role played by MontyLingua and its component in these research. Section 4 discusses some trends observed in these research. However, it is not the aim of this paper to describe MontyLingua itself or the works using it, at the source code level.

## 2. Distinctive Feature of MontyLingua

The distinctive feature of MontyLingua is the coverage for all aspects of English text processing from raw input text to semantic meanings and summary generation, yet each component in MontyLingua is loosely-coupled to each other at the architectural and code level. This had enabled MontyLingua to be used in 3 different contexts: (1) as a suite of tools for processing text to semantic meaning and summary generation; (2) decouple each component of MontyLingua for individual use; (3) using MontyLingua as a baseline system and substituting components to cater to specific applications. The end result of (2) and (3) may be the same but the approaches are philosophically different. The rest of this section will focus on the individual components making up MontyLingua and how (2) and (3) can be fulfilled.

MontyLingua consists of six components: MontyTokenizer, MontyTagger, MontyLemmatiser, MontyREChunker, MontyExtractor, and MontyNLGenerator. MontyTokenizer, which is sensitive to common abbreviations, separates the input English text into constituent words and punctuations. Common contractions are resolved into their un-contracted form. For example, "you're" is resolved to "you are". MontyTagger is a Penn Treebank Tag Set (Marcus et al., 1993) part-of-speech (POS) tagger based on Brill tagger (Brill, 1995) and enriched with commonsense in the form of contextual rules. MontyLemmatiser strips any inflectional morphology from each word. That is, verbs are reduced to infinite form and nouns to singular form. MontyREChunker reads the POS sequence and identifies semantic phrases (adjective, noun, verb, prepositional) using a series of Regular Expressions. MontyExtractor extracts phrases and subject-verb-object triplets from the chunked text. Lastly, MontyNLGenerator uses the output of MontyExtractor to generate text summaries.

At code level, each component resides in a file and is standalone. This feature enables each of the six components to be used individually. In some of the research articles reviewed in Section 4 below, MontyTagger was used on its own. On the other hand, it also means that each of the six components can be easily substituted to cater to specific applications. The simplest way to do this is to modify the *jist* method in the class MontyLingua (file: MontyLingua.py) as follows: The *jist* method illustrates the end-to-end process of MontyLingua.

```
def jist(self,text):
    sentences = self.split_sentences(text)
    tokenized = map(self.tokenize,sentences)
    tagged = map(self.tag_tokenized,tokenized)
    chunked = map(self.chunk_tagged,tagged)
    #print "CHUNKED: " + string.join(chunked,'\n     ')
    extracted = map(self.extract_info,chunked)
    return extracted
```

The input text is tokenized, tagged, chunked by MontyTokenizer, MontyTagger, and MontyREChunker respectively before phrase and subject-verb-object triplets are extracted by MontyExtractor. Substituting each of these component is little more than re-directing the execution to the substituted component and back.

## 3. Anthology of Applications Utilizing MontyLingua

Six research articles were retrieved from ACM Digital Library using "montylingua" as the search term. A search using Google (search term: +montylingua +.pdf) added another 13 to the list; consisting of 1 doctoral dissertation, 1 masters dissertation, 2 technical reports, and 9 articles. This section will briefly describe the role of MontyLingua in each of these 19 publications published between October 2004 and August 2006 in chronological order.

### 3.1. Chandrasekaran's Adaptive Multimodal Language Acquisition

(Chandrasekaran, 2004) attempted to develop a language acquisition system through multimodal input. The system tries to initiate a dialog with the users to learn nouns, verbs, or adjectives. Text input were POS tagged by MontyTagger to identify nouns, verbs, or adjectives.

### 3.2. ATHENS

ATHENS system (Skillicorn and Vats, 2004), developed in Queen's University, Canada, is a web-mining tool for information discovery. A case study on extracting knowledge on terrorism was presented. The authors extracted 9 clusters of information which summarized the events as of September 12, 2001 using the search terms "al Qaeda" and "bin Laden". After retrieving a list of web-pages through Google WebAPI, MontyTagger was used to generate a list of nouns, which was then filtered for a list of discriminatory nouns by comparison to the relative frequency in British National Corpus (www.natcorp.ox.ac.uk). A page-page Jaccard similarity matrix (Bradeen and Havey, 1995) was computed using the frequencies of discriminatory nouns on each page which considered multiple search terms (2 search terms in this case). Finally, a 2-pass clustering was performed – first on the entire set of retrieved web-pages, followed by clustering within each of the top level clusters. A list of descriptive nouns were generated for each cluster. Iterative search can be done using the list of descriptors for each cluster.

### 3.3. HyperPipes

Eisenstein and Davis (2004) attempted to develop a human gesture classifier, *HyperPipes*, into 4 categories (deictic, action, other, unknown) using only linguistics information. A set of manually classified gestures with the corresponding transcribed speech were extracted from 9 persons (not physics or mechanically trained) describing 3 objects: a latchbox, a piston, and a pinball machine. MontyLingua was used for POS tagging and stemming of the transcribed speech. A number of features were extracted from MontyLingua-processed text, including unigrams, bigrams and trigrams. Comparing a baseline classification where all gestures are deictic (48.7% accurate), HyperPipes achieved an accuracy of 66%. This was compared to Naïve Bayes (59%), C4.5 (56%) and SVM (56%). This was also compared to manual classification with only audio information, that is, humans listening to the speech without watching the video footage, which only achieved 45% accuracy.

### 3.4. Udani et. al.'s Noun Sense Induction

Word sense induction refers to inferring contextual senses of an ambiguous word (words with multiple meanings) which is a crucial aspect of text understanding. Udani et. al. (2005) attempted to advance this field by bootstrapping on the the large body of contextual information available online for sense induction of nouns. MontyLingua was used to tag and stem the first 500 research result titles and snippets from Google for clustering. The system was evaluated on 5 terms and demonstrated 85.7% accuracy in noun sense induction as compared to the random chance of 31.6% accuracy.

### 3.5. MontyTagger as a Teaching Tool
Light et. al. (2005) observed increasing numbers of non-computer science student interested in learning about natural language processing. However, these students had difficulty in understanding programming and Unix to use computational linguistics tools effectively. Light et. al. (2005) constructed a web-based interface to nine computational linguistic tools, including MontyTagger.

### 3.6. Text Processing of Economics Literature
Nee Jan van Eck's masters dissertation at the Econometric Institute of Erasmus University Rotterdam focused on text processing of economics literature for the purpose of extracting economics-relevant terms and presenting it as a concept map linking these terms (van Eck, 2005, van Eck and van den Berg, 2005). MontyLingua was used to tokenize, POS tag, and stem economics literature prior to linguistics and statistical filtering for relevant terms.

### 3.7. Metafor
Metafor was developed as a structure generation tool to convert everyday English language into Python codes (Liu and Lieberman, 2005), which is a common task for programmers who need to implement requirements into systems. MontyLingua was used to process input text into subject-verb-object(s) triplets which were anaphorically dereferenced using ConceptNet (Liu and Singh, 2004). Programmatic entities forming the core generated codes were performed in three parts. Firstly, a set of semantic recognizers were used over the subject-verb-object(s) triplets to identify code structures, such as lists, quotes, and if-else structures. Secondly, actions or changes to the extracted code structures were identified which would be used to form the class functions. Lastly, the context of the actions were identified. That is, which actions affect which objects. These programmatic entities were then used to generate Python codes. Although it is not likely that the generated Python code is executable, Metafor is likely to be adopted as a brainstorming tool according to a case study done by the authors (Liu and Lieberman, 2005).

### 3.8. Richardson and Fox's Concept Map Based Cross-Language Resource Learning
Concept map was described by Joseph Novak as "*graphical representations of knowledge that are comprised of concepts and the relationships between them*" (Novak and Gowin, 1984) which had been shown to facilitate a student's learning process (McNaught and Kennedy, 1997). Richardson and Fox (2005) examined the role of concept maps as a cross-language learning resource by giving a set of articles written in Spanish and their English translations to a control group of student, whereas the experimental group received the same materials as the control group supplemented with concept maps produced by domain experts. The experimental group performed significantly better than the control suggesting the advantage of having a concept map. MontyTagger was used to extract nouns which were subsequently used to form the nodes on the concept automatically in further experiments but the authors did not evaluate the differences in the nodes of the concept maps produced by domain experts and that of MontyTagger.

### 3.9. QABLe
QABLe (Question-Answering Behavior Learner) used prior learning and problem solving strategies (Tadepalli and Natarajan, 1996) in text understanding for question and answer (Grois and Wilkins, 2005b, Grois and Wilkins, 2005a). MontyTagger was used for both processing of text, which was to be understood, and the questions. A prior system, Deep Read (Hirschman et al., 1999), was evaluated using Remedia Corpus (a collection of 115 children's stories provided by Remedia Publications). Using the same corpus, QABLe achieved 48% accuracy, compared to 36% by Deep Read (Grois and Wilkins, 2005b, Grois and Wilkins, 2005a).

### 3.10. Arizona State University BioAI group in TREC 2005
The Text Retrieval Conference (TREC) Genomic Track 2005 is an ad-hoc document retrieval task in 5 different areas of 10 instances each. The Arizona State University BioAI group (Yu et al., 2005) chose to use Apache Lucene (lucene.apache.org) to retrieve abstracts from PubMed, which were POS tagged using MontyTagger and anaphorically resolved. Facts from the processed abstracts were extracted by template matching. Evaluations by TREC were based on the top 10 and 100 retrieved abstracts respectively. Yu et. al. (Yu et al., 2005) achieved 27% precision and 11% precision on the top 10 and 100 abstracts respectively.

### 3.11. SkillSum
Reiter and Dale said that "the goal of many NLG [natural language generation] systems is to produce documents which are as similar as possible to documents produced by human experts" (Reiter and Dale, 2000). One of the difficulties is to decide what goes into the generated document, the context selection rules, and it is also known that corpora of expert-written text may not form the gold standards as expert may disagree or vary in opinions (Reiter and Sripada, 2002). From a set of skills test results and authored evaluations, SkillSum attempted to derive context rules (Williams and Reiter, 2005). MontyLingua was used to parse authored evaluations to identify message types (Geldof, 2003), followed by Rhetorical Structure Theory analysis. A trial by the authors suggested that users preferred SkillSum's report over basic numerical test scores (Reiter and Dale, 2000).

### 3.12. Kennedy, Natsev and Chang's Query Class Induction for Multimodal Video Search
One of the more sophisticated forms of search techniques is multimodal search which assumes the set of items to be searched takes on different roles and specific search techniques, when applied, could improve overall retrieval performance. For example, a video clip in a collection could be searched by title and subject classification (metadata), qualities of image or contents of image (visual cues), dialogue or speech (audio cues), and subtitles (text). In multimodal search, an important aspect is to be able to classify the search queries and studies in multimodal video retrieval had used pre-defined classes (Chua et al., 2004, Yan et al., 2004). Kennedy, Natsev and Chang proposed a framework for multimodal search without prior need to define query classes by semantic analysis of the input query (Kennedy et al., 2005). MontyLingua was used for POS tagging and stemming of the input query before constructing it into an OKAPI query (Robertson et al., 1992). An improvement of 18% was realized over using pre-defined query classes (Chua et al., 2004, Yan et al., 2004) by evaluating using TRECVID 2004 (Robertson et al., 1992).

### 3.13. Memsworldonline
Memsworldonline (Zhang et al., 2006a) was developed for information retrieval in domain-specific digital libraries on microelectromechanical systems by using a combination of Formal Concept Analysis (Priss, 1996) and information anchors. Information anchors are common concepts in the field which allowed for examination into community dynamics (Troy et al., 2006) or emerging trends (Kontostathis et al., 2003). For example, this paper is an information anchor for MontyLingua (topic area). Other possible anchors are authors (related areas of expertise) and institutions (research directions). Information anchors essentially consists of keywords, key phrases, metadata, and inter-document relationships. MontyLingua was used in Memsworldonline to extract nouns, noun phrases, and sub-phrases in documents as one of the means to derive information anchors. These information anchors formed an ontology to classify documents.

### 3.14. PEPURS
With increasing use of digital libraries comes the problem of author ambiguity (Torvik et al., 2005), as author names could be written in various forms of initials and more than one published authors may share the same initial. PEPURS attempted to advance the field of author name clarification by analyzing author's websites for publication records and segmenting these records into appropriate data fields (Zhang et al., 2006b). Each publication record is tagged twice, once by a purpose-built tagger, and by MontyTagger. These were then used as input for B-classifier and P-classifier running in parallel to segment the publication records before merging the results from the classifiers using a decision tree (Mitchell, 1997). The three classifiers ran as a stacked generalization procedure (Wolpert, 1992).

### 3.15. Automatic Construction of Domain-Specific Concept Structures
Libo Chen's doctoral dissertation at Technischen Universitat Darmstadt focused on automatic

construction of domain-specific concept structures (Chen, 2006) in response to the problem of vocabulary mismatch in web search (Blair, 1986, Furnas et al., 1987) by constructing domain-specific concepts and linking these terms. MontyTagger was used to POS tag web-pages prior to concept extraction.

### 3.16. Red Opal
Feature selection of online product reviews is an important aspect of online shopping (Liu et al., 2005). Red Opal (Scaffidi, 2006) used a probability-based algorithm in feature selection, and comparing that to a support-based algorithm (Liu et al., 2005). MontyLingua was used for POS tagging and stemming of online product reviews before processing by each of the two algorithms for feature selection. The speed of MontyLingua's POS tagging and stemming averaged at 301 milliseconds per review, with the fastest being 250 milliseconds, on a single 3GHz Pentium 4 processor with 1GB of RAM, running Windows XP Professional SP 2 and Sun's J2RE 1.4.2 with 250MB heap size.

### 3.17. Hsu and Chen's Commonsense Query Expansion for Image Retrieval
Hsu and Chen (2006) investigated the usefulness of commonsense knowledge in image retrieval which had been used previously in query expansion (Liu and Lieberman, 2002). MontyLingua was used for POS tagging and stemming of the initial query before commonsense query expansion by ConceptNet (Liu and Singh, 2004). From the evaluation results using the ImageCLEF 2005 test collection (Clough et al., 2005), the authors concluded that introducing commonsense knowledge into the retrieval task is suitable for precision-oriented tasks (Hsu and Chen, 2006).

## 4. Discussion
MontyLingua was released in 2004 (Liu, 2004) and was described in October 2004 with ConceptNet (Liu and Singh, 2004). In the same paper, 15 applications of ConceptNet were featured. Since then, the state-of-the-art use of either ConceptNet or MontyLingua and roles played by MontyLingua and its components in recent research work had not been reviewed. This paper aims to review the use of and roles played by MontyLingua and its components in research work published between October 2004 to August 2006.

Of the 17 research reports reviewed, all had used MontyTagger for POS tagging, 8 of them had used MontyLemmatiser for stemming, and only 2 (Metafor and Memsworldonline) had used MontyREChunker and MontyExtractor. None of the reviewed work seems to have used MontyNLGenerator for text summarization.

An interesting observation is the use of MontyTagger in a wide context, such as web-pages (Skillicorn and Vats, 2004, Udani et al., 2005), transcribed human speech (Eisenstein and Davis, 2004), economics papers (van Eck, 2005, van Eck and van den Berg, 2005), and biomedical papers (Yu et al., 2005), despite the fact that MontyTagger was generically trained using Wall Street Journal corpus. This might suggest that MontyTagger could be used in various context, which is reflected in daily life where a non-legally trained person might still be able to read legal text intelligently despite some inability to grasp the total meaning as appear to a legally trained person. However, it had been shown that a generically trained POS tagger will perform inadequately on domain-specialized text, such as biomedical literature (Tateisi and Tsuji, 2004). In spite of this, MontyTagger had been used in specialized sitting (van Eck, 2005, van Eck and van den Berg, 2005, Yu et al., 2005) which might suggest that the numerical measurement of POS tagging accuracy may not correlate with the *"functional"* POS tagging accuracy. For example, the word "book" can be tagged as "noun, base form" (NN) or "noun, singular form" (NNS) but may be treated as an error when calculating POS tagging accuracy as the quotient between the number of correctly tagged tokens and the total number of tokens.

Only 2 of the systems had used MontyREChunker and MontyExtractor. Metafor had used them to gain semantic understanding of daily written language while Memsworldonline used them to process domain-specific text. Despite a small sample size of 2, a supportive case could be made for the use of MontyREChunker and MontyExtractor in both generic text (Liu and Lieberman, 2005) and domain-specific text (Zhang et al., 2006a).

In this review, we did not observe any applications of MontyNLGenerator. However, it is likely that text summary may have a role in future in managing the current trend of information overload. It is plausible that future research will place greater emphasis on summary generation of domain-specific libraries as a whole or in a time-striated fashion, as an extension of Memsworldonline. Web search could use natural language generation techniques to summarize the results on-the-fly. Natural language generation could extend Metafor (Liu and Lieberman, 2005) to include automated generation source code documentation. This could then be used to identify code architectures and algorithms which is one of the problems in program optimization by algorithm replacement (Metzger and Wen, 2000).

In summary, we had reviewed 19 articles published between October 2004 and August 2006 for the roles played by MontyLingua or its components in these studies, thereby updating the state-of-the-art utility of MontyLingua. We had observed a diversified use of MontyLingua in many different areas, both generic and domain-specific. Although the use of the text summarizing component had not been observed, we are optimistic that it will have a crucial role in managing the current trend of information overload in future research.

## 5.     Acknowledgement

The author will like to thank the reviewers for their invaluable comments.

## 6.     References

BLAIR, D. C. (1986) Indetermincy in the subject access to documents. *Information Processing and Management,* 22**,** 229-241.
BRADEEN, J. M. & HAVEY, M. J. (1995) Restriction Fragment Length Polymorphisms Reveal Considerable Nuclear Divergence within a Well-Supported Maternal Clade in Allium Section Cepa (Alliaceae). *American Journal of Botany,* 82, 1455-1462.
BRILL, E. (1995) Transformation-based error-driven learning and natural language processing: a case study in part of speech tagging. *Computational Linguistics,* 21**,** 543-565.
CHANDRASEKARAN, R. (2004) Using language structure for adaptive multimodal language acquisition. *Sixth International Conference on Multimodal Interfaces (ICMI'04).* State College, Pennsylvania, USA, ACM Press.
CHEN, L. (2006) Automatic construction of domain-specific concept structures. Technischen Universitat Darmstadt.
CHUA, T. S., NEO, S. Y., WANG, K. Y., SHI, R., ZHAO, M. & XU, H. (2004) TRECVID 2004 search and feature extraction task by NUS PRIS. *TRECVID 2004 Workshop.*
CLOUGH, P., MULLER, H., DESELAERS, T., GRUBINGER, M., LEHMANN, T. M., JENSEN, J. & HERSH, W. (2005) The CLEF 2005 cross-language image retrieval track. *2005 Cross Language Evaluation Forum.*
COHEN, W. W. (2004) Minorthird: Methods for Identifying Names and Ontological Relations in Text using Heuristics for Inducing Regularities from Data, http://minorthird.sourceforge.net
CUNNINGHAM, H. (2000) Software Architecture for Language Engineering. *Department of Computer Science.* University of Sheffield.
EISENSTEIN, J. & DAVIS, R. (2004) Visual and linguistic information in gesture classification. *6th international conference on Multimodal interfaces.* State College, PA, USA, ACM Press.
FURNAS, G. W., LANDAUER, T. K., GOMEZ, L. M. & DUMAIS, S. T. (1987) The vocabulary problem in human-system communication. *Communications of the ACM,* 30**,** 964-971.
GELDOF, S. (2003) Corpus analysis for NLG. IN REITER, E., HORACEK, H. & DEEMTER, K.

V. (Eds.) *9th European Workshop on NLG.*
GROIS, E. & WILKINS, D. (2005a) Learning strategies for story comprehension: a reinforcement learning approach. *22nd International Conference on Machine Learning.* Bonn, Germany.
GROIS, E. & WILKINS, D. C. (2005b) Learning Strategies for Open-Domain Natural Language Question Answering. *International Joint Conference on Artificial Intelligence 2005 (IJCAI05).*
HIRSCHMAN, L., LIGHT, M. & BURGER, J. (1999) Deep Read: A reading comprehension system. *Annual Meeting of the Association of Computational Linguistics 99.*
HSU, M.-H. & CHEN, H.-H. (2006) Information retrieval with commonsense knowledge. *29th annual international ACM SIGIR conference on Research and development in information retrieval.* Seattle, Washington, USA, ACM Press.
KENNEDY, L. S., NATSEY, A. & CHANG, S.-F. (2005) Automatic discovery of query-class-dependent models for multimodal search. *13th annual ACM international conference on Multimedia.* Singapore, ACM Press.
KONTOSTATHIS, A., GALITSKY, L. M., POTTENGER, W. M., ROY, S. & PHELPS, D. J. (2003) A survey of emerging trend detection in textual data mining. IN BERRY, M. (Ed.) *A comprehensive survey of text mining.* Springer-Verlag.
LIGHT, M., ARENS, R. & LU, X. (2005) Web-based interfaces for natural language processing tools. *43rd Annual Meeting of the Association for Computational Linguistics - Effective Tools and Methodologies for Teaching Natural Language Processing and Computational Linguistics Workshop.*
LIU, B., HU, M. & CHENG, J. (2005) Opinion Observer: Analyzing and comparing opinions on the web. *14th International Conference on World Wide Web (WWW'05).* ACM Press.
LIU, H. (2004) MontyLingua: An end-to-end natural language processor with common sense.
LIU, H. & LIEBERMAN, H. (2002) Robust photo retrieval using world semantics. *LREC 2002 Workshop on Creating and Using Semantics for Information Retrieval and Filtering.* Canary Islands.
LIU, H. & LIEBERMAN, H. (2005) Metafor: visualizing stories as code. *10th International Conference on Intelligent User Interfaces.* San Diego, California, USA.
LIU, H. & SINGH, P. (2004) ConceptNet: A Practical Commonsense Reasoning Toolkit. *BT Technology Journal,* 22**,** 211-226.
LOPER, E. & BIRD, S. (2002) NLTK: The natural language toolkit. *ACL Workshop on Effective Tools and Methodologies for Teaching Natural Language Processing and Computational Linguistics.* Philadelphia, Association for Computational Linguistics.
MARCUS, M. P., SANTORINI, B. & MARCINKIEWICZ, M. A. (1993) Building a large annotated corpus of English: the Penn Treebank. *Computational Linguistics,* 19**,** 313-330.
MCNAUGHT, C. & KENNEDY, D. (1997) Use of concept mapping in the design of learning tools for interactive multimedia. *Journal of Interactive Learning Research,* 8**,** 389-406.
METZGER, R. & WEN, Z. (2000) *Automatic algorithm recognition and replacement: A new approach to program optimization*, The MIT Press.
MITCHELL, T. (1997) *Machine Learning*, McGraw Hill.
NOVAK, J. D. & GOWIN, D. B. (1984) *Learning how to learn,* Cambridge, UK, Cambridge University Press.
PRISS, U. (1996) Formal concept analysis in information science. *Annual Review of Information Science and Technology,* 40.
REITER, E. & DALE, R. (2000) *Building natural language generation systems,* Cambridge, Cambridge University Press.
REITER, E. & SRIPADA, S. (2002) Should corpora text be gold standards for NLG? *International Conference of Natural Language Generation 2002 (INLG02).*
RICHARDSON, R. & FOX, E. A. (2005) Using concept maps in digital libraries as a cross-language resource discovery tool. *5th ACM/IEEE-CS Joint Conference on Digital libraries.* Denver, CO, USA, ACM Press.
ROBERTSON, S. E., WALKER, S., HANCOCK-BEAULIEU, M., GULL, A. & LAU, M. (1992) Okapi at TREC4. *Text Retrieval Conference 1992.*

SCAFFIDI, C. (2006) Application of a probability-based algorithm to extraction of product features from online reviews. Pittsburg, PA, USA, Carnegie Mellon University.

SKILLICORN, D. B. & VATS, N. (2004) Novel information discovery for intelligence and counterterrorism. Kingston, Ontario, Canada, School of Computing, Queen's University.

TADEPALLI, P. & NATARAJAN, B. (1996) A formal framweork for speedup learning from problems and solutions. *Journal of Artificial Intelligence Research,* 4**,** 445-475.

TATEISI, Y. & TSUJI, J. I. (2004) Part-of-Speech Annotation of Biology Research Abstracts. *4th International Conference on Language Resource and Evaluation (LREC2004).*

TORVIK, V. I., WEEBER, M., SWANSON, D. R. & SMALHEISER, N. R. (2005) A probabilistic similarty metric for Medline records: a model for author name disambiguation. *Journal of the American Society for Information Science and Technology,* 56**,** 140-158.

TROY, A. D., ZHANG, G. Q. & MEHREGANY, M. (2006) Evolution of the Hilton Head Workshop research community. *Education Digest of the 2006 Solid-State Sensor and Actuator Workshop.*

UDANI, G., DAVE, S., DAVIS, A. & SIBLEY, T. (2005) Noun sense induction using web search results. *28th Annual International ACM SIGIR Conference on Research and Development in Information Retrieval.* ACM Press.

VAN ECK, N. J. (2005) Towards automatic knowledge discovery from scientific literature. *Econometric Institute, Faculty of Economics.* Rotterdam, Erasmus University.

VAN ECK, N. J. & VAN DEN BERG, J. (2005) A novel algorithm for visualizing concept associations. *16th International Workshop on Database and Expert System Applications (DEXA'05).*

WILLIAMS, S. & REITER, E. (2005) Deriving Content Selection Rules from a Corpus of Non-naturally Occurring Documents for a Novel NLG Application. IN BELZ, A. & VARGES, S. (Eds.) *Corpus Linguistics 2005 Workshop on Using Corpora for Natural Language Generation.* Birmingham, UK.

WOLPERT, D. (1992) Stacked generalization. *Neural Networks,* 5**,** 241-259.

YAN, R., YANG, J. & HAUPTMANN, A. G. (2004) Learning query-class dependent weights in automatic video retrieval. *ACM Multimedia 2004.* ACM Press.

YU, L., AHMED, S. T., GONZALEZ, G., LOGSDON, B., NAKAMURA, M., NIKKILA, S., SHAH, K., TARI, L., WENDT, R., ZEIGLER, A. & BARAL, C. (2005) Genomic information retrieval through selective extraction and tagging by the ASU-BioAI group. *14th Text Retrieval Conference (TREC2005).*

ZHANG, G.-Q., TROY, A. D. & BOURGOIN, K. (2006a) Bootstrapping ontology learning for information retrieval using formal concept analysis and information anchors. *14th International Conference on Conceptual Structures.* Aalborg, Denmark.

ZHANG, W., YU, C., SMALHEISER, N. & TORVIK, V. (2006b) Segmentation of publication records of authors from the web. *22nd IEEE International Conference on Data Engineering (ICDE'06).* Atlanta, Georgia, IEEE Press.